\documentclass[conference]{IEEEtran}
\IEEEoverridecommandlockouts
\usepackage{cite}
\usepackage{amsmath,amssymb,amsfonts}
\usepackage{algorithmic}
\usepackage{graphicx}
\usepackage{textcomp}
\usepackage{xcolor}
\usepackage{url}
\usepackage{placeins}
\usepackage{hyperref}
\usepackage{float}
\usepackage{tabularx,colortbl}
\usepackage{ifthen}
\usepackage{caption,subcaption}
\usepackage{amsmath}
\usepackage{booktabs}
\renewcommand{\thetable}{\Roman{table}}
\usepackage{cleveref}

\usepackage{enumitem}
\usepackage{stfloats}
\usepackage{wrapfig}
\usepackage{verbatim}
\usepackage{tabularx}
\usepackage{booktabs}
\usepackage{multirow}
\usepackage{caption}
\usepackage{flushend}
\usepackage{amsmath}

\usepackage[a4paper, total={184mm,239mm}]{geometry}
\def\BibTeX{{\rm B\kern-.05em{\sc i\kern-.025em b}\kern-.08em
    T\kern-.1667em\lower.7ex\hbox{E}\kern-.125emX}}
\def\BibTeX{{\rm B\kern-.05em{\sc i\kern-.025em b}\kern-.08em
    T\kern-.1667em\lower.7ex\hbox{E}\kern-.125emX}}
\begin{document}

\title{Intelligent Sensing-to-Action for Robust Autonomy at the Edge: \textit{Opportunities and Challenges}}

\author{Amit Ranjan Trivedi$^1$, Sina Tayebati$^1$, Hemant Kumawat$^2$, Nastaran Darabi$^1$, Divake Kumar$^1$, \\ Adarsh Kumar Kosta$^4$, Yeshwanth Venkatesha$^3$, Dinithi Jayasuriya$^1$, Nethmi Jayasinghe$^1$, Priyadarshini Panda$^3$, \\ Saibal Mukhopadhyay$^2$, and Kaushik Roy$^4$ \vspace{2mm} \\
$^1$University of Illinois at Chicago, $^2$Georgia Institute of Technology, $^3$Yale University, $^4$Purdue University}
\maketitle

\begin{abstract}
Autonomous edge computing in robotics, smart cities, and autonomous vehicles relies on the seamless integration of sensing, processing, and actuation for real-time decision-making in dynamic environments. At its core is the \textit{sensing-to-action loop}, which iteratively aligns sensor inputs with computational models to drive adaptive control strategies. These loops can adapt to hyper-local conditions, enhancing resource efficiency and responsiveness, but also face challenges such as resource constraints, synchronization delays in multi-modal data fusion, and the risk of cascading errors in feedback loops. This article explores how proactive, context-aware sensing-to-action and action-to-sensing adaptations can enhance efficiency by dynamically adjusting sensing and computation based on task demands, such as sensing a very limited part of the environment and predicting the rest. By guiding sensing through control actions, action-to-sensing pathways can improve task relevance and resource use, but they also require robust monitoring to prevent cascading errors and maintain reliability. Multi-agent sensing-action loops further extend these capabilities through coordinated sensing and actions across distributed agents, optimizing resource use via collaboration. Additionally, neuromorphic computing, inspired by biological systems, provides an efficient framework for spike-based, event-driven processing that conserves energy, reduces latency, and supports hierarchical control--making it ideal for multi-agent optimization. This article highlights the importance of end-to-end co-design strategies that align algorithmic models with hardware and environmental dynamics, improve cross-layer interdependencies to improve throughput, precision, and adaptability for energy-efficient edge autonomy in complex environments.
\end{abstract}


\section{Introduction}
Autonomous edge computing in domains such as robotics, smart cities, and autonomous vehicles relies on the seamless integration of sensing, processing, and actuation to enable real-time decision-making in dynamic environments. Central to this integration is the \textit{sensing-to-action loop}, which iteratively aligns sensor inputs with computational models to drive adaptive control strategies. Unlike centralized systems that rely on static, generalized models, sensing-to-action loops can seamlessly adapt to hyper-local conditions, such as environmental variations, sensor and processor health, and task-specific priorities, thereby enabling optimized resource allocation, reduce communication latency, and faster responsiveness, and minimizing dependencies on external networks.

However, the sophisticated manipulability of sensing-to-action loops at the edge also presents significant challenges. Unlike predominant deep learning pipelines that primarily focus on feed-forward sensing-to-insight, where latency primarily impacts inference speed, delays in \textit{cyclical} sensing-to-action loops risk propagating outdated environmental states, thereby significantly degrading decision accuracy. Therefore, while sensing-to-insight pathways are more amenable to cloud-based centralized computations, such as under batch processing, edge systems must perform continuous, localized sensing-action loop computations for real-time hypothesis testing and action refinement, making them highly sensitive to resource constraints. Additionally, improving system observability often requires high-fidelity, multi-modal sensors, which can be resource-intensive and impractical due to constraints on power, bandwidth, and form factor. The fusion of heterogeneous, high-bandwidth sensor streams further introduces synchronization delays, communication overhead, and latency issues.

In this perspective article, we closely examine such challenges in closed-loop sensing-to-action mechanisms at the edge, while also highlighting untapped and emerging opportunities to enable precise, low-latency control with minimal resource requirements. By shifting from reactive sensing-to-insight pipelines to proactive, context-aware sensing-action loop adaptations, we explore strategies essential for autonomous systems operating in complex environments. Unlike traditional machine learning pipelines, the bi-directional information flow in sensing-action loops offers unprecedented co-optimization opportunities, allowing systems to dynamically adjust sensing and computation based on task demands and environmental context. For example, tasks less sensitive to sensor noise or feature reduction can be executed under lower signal-to-noise ratios or reduced precision, conserving resources.

Likewise, sensing-to-action loops can be fine-tuned based on scene-specific dynamics, enabling systems to allocate resources more efficiently by adjusting sensor refresh rates, resolution, or modality usage in response to environmental changes. For example, environmental monitoring sensors can reduce their sampling rates during stable periods and increase them during sudden events, such as pollutant surges. Similarly, autonomous systems can deprioritize redundant sensor streams during low-risk tasks while enhancing accuracy for high-stakes operations. These loops also support hierarchical control, where low-level actions -- such as adjusting sensor thresholds -- complement higher-level planning decisions, enabling efficient distribution of computational effort. By leveraging such interplay between sensing and action, sensing-to-action loops therefore open new pathways for adaptive, energy-efficient edge autonomy.

Moreover, we discuss how beyond conventional digital and analog processing, spike-based representations in neuromorphic computing domain provide a natural framework for sensing-to-action loop optimization. Unlike clock-driven architectures, neuromorphic systems use event-driven, asynchronous computations, where spikes trigger processing only for relevant sensory events, reducing latency and energy use. This sparse encoding dynamically adjusts computational loads inherently based on activity levels. Neuromorphic architectures also excel in hierarchical and distributed processing, enabling local spiking circuits to handle low-level actions (e.g., sensor threshold adjustments) while coordinating high-level planning through more complex pathways. Their inherent parallelism and decentralized design make them ideal for multi-agent optimization, where collaborative sensing and resource allocation are crucial. 

Finally, multi-agent sensing-action loops further extend the potential of sensing-to-action loops through distributed optimization. By understanding the interactions between sensing and actuation groups, agents can dynamically adjust resource allocation based on collaboration and task requirements. Through information sharing and coordinated actions, agents can optimize sensing and processing resources across the network. For example, one agent can reduce its sensing load if another has superior coverage or access to relevant data, improving overall system efficiency. This distributed coordination makes sensing-to-action loops suited for edge autonomy.

Towards exploring such opportunities for sensing-action loop optimization, a central focus of the paper is to underscore the importance of \textit{end-to-end co-design strategies} that align algorithmic models with hardware constraints and environmental dynamics to improve latency, energy efficiency, and robustness. Unlike modular optimizations that only address individual components in isolation, end-to-end approaches can leverage cross-layer interdependencies, unlocking unprecedented gains in throughput, precision, and resource allocation. 

\begin{figure}[t!]
    \centering
    \includegraphics[width=\linewidth]{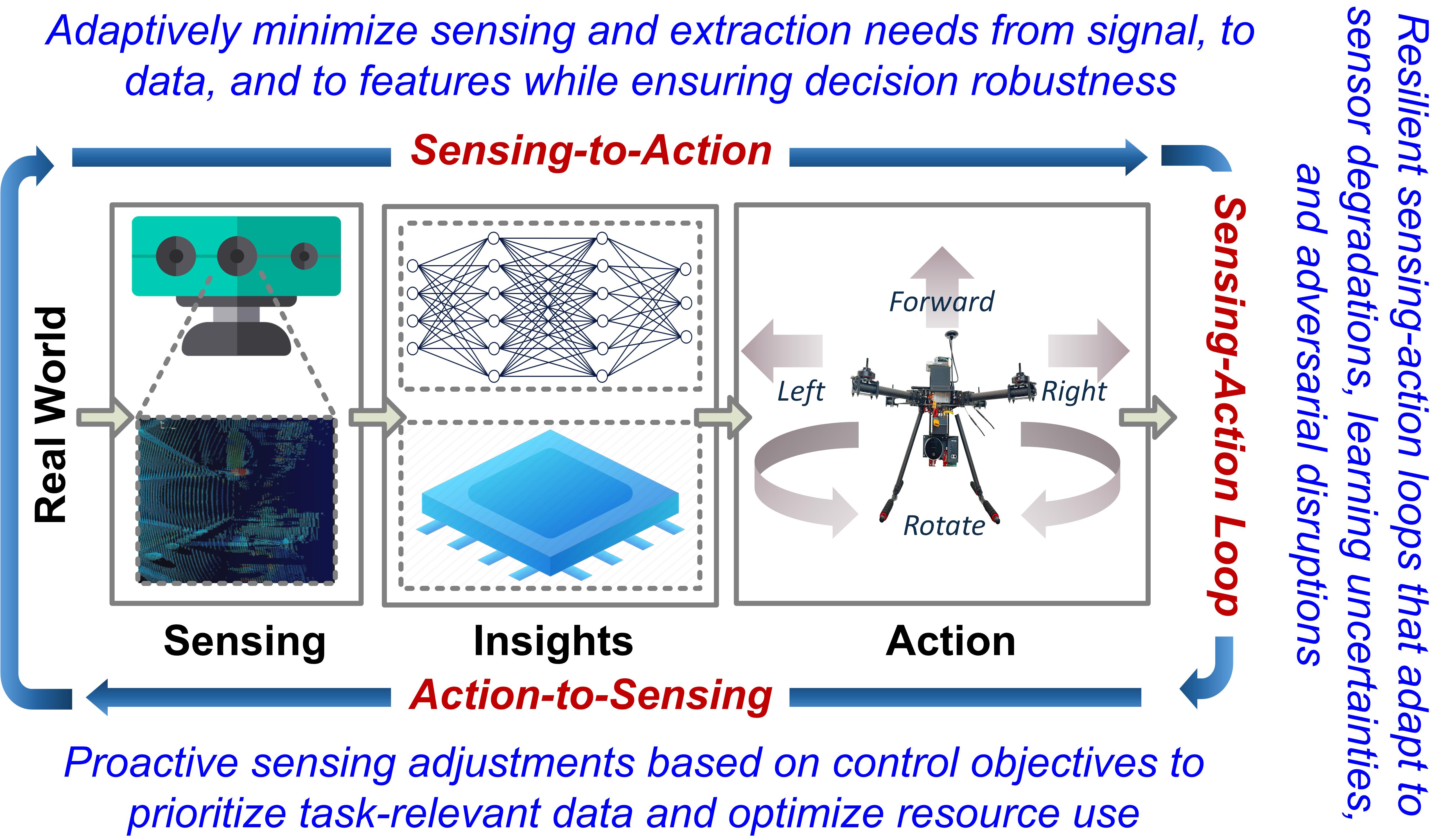}
\caption{\textbf{Opportunities for Intelligent Sensing-to-Action:} In \textit{sensing-to-action} loops, significant gains can be achieved by selectively sensing critical environmental regions while \textit{predicting} less critical areas based on training data. This frugal sensing strategy is especially beneficial for resource-intensive modalities, such as LiDAR, enhancing task accuracy without unnecessary overhead. Similarly, \textit{action-to-sensing} optimizations can adjust control variables to opportunistically reduce sensing demands based on task relevance. While these frameworks improve loop efficiency, ensuring reliability requires robust and computationally efficient monitors to continuously assess fidelity and support aggressive optimizations. In \textit{multi-agent sensing-action loops}, agents can collaborate by sharing sensing tasks or complementing each other's sensing capabilities. Moreover, emerging paradigms, such as \textit{neuromorphic sensing-action loops}, offer unified frameworks by adapting sensing and processing rates based on event dynamics, enabling seamless sensing and control.}
    \label{fig:opportunities}
\end{figure}

\section{Intelligent Sensing-to-Action}
In Fig.\ref{fig:opportunities}, at a high level, sensing-to-action loops at the edge can be deconstructed into three primary components: the sensing module, the learning module, and the actuation module. This process begins with the environment generating stimuli that are captured by sensors and converted into data streams for downstream processing. The sensing block often handles multiple modalities—such as vision, sound, and environmental readings—that must be fused and pre-processed to extract meaningful features. These features are then passed to the processing pipeline, where machine learning or decision-making models predict control actions. These actions, executed by actuators, influence the environment, completing the loop as the updated environment feeds into the next sensing stage. 

While these loops enable low-latency decision-making, they also introduce unique challenges due to the constrained nature of edge devices. Limited computational resources, memory, and energy impose strict trade-offs between model complexity and real-time performance. High-fidelity sensing, though necessary for reliable feature extraction, can quickly overwhelm edge hardware due to power and bandwidth constraints. Additionally, fusing heterogeneous data streams can lead to synchronization delays and communication overhead, further complicating real-time feedback. The cyclical nature of the loop also amplifies sensitivity to outdated or noisy data, as errors can propagate and compound, degrading downstream decisions.

\begin{figure*}[t]
  \centering
\includegraphics[width=0.92\linewidth]{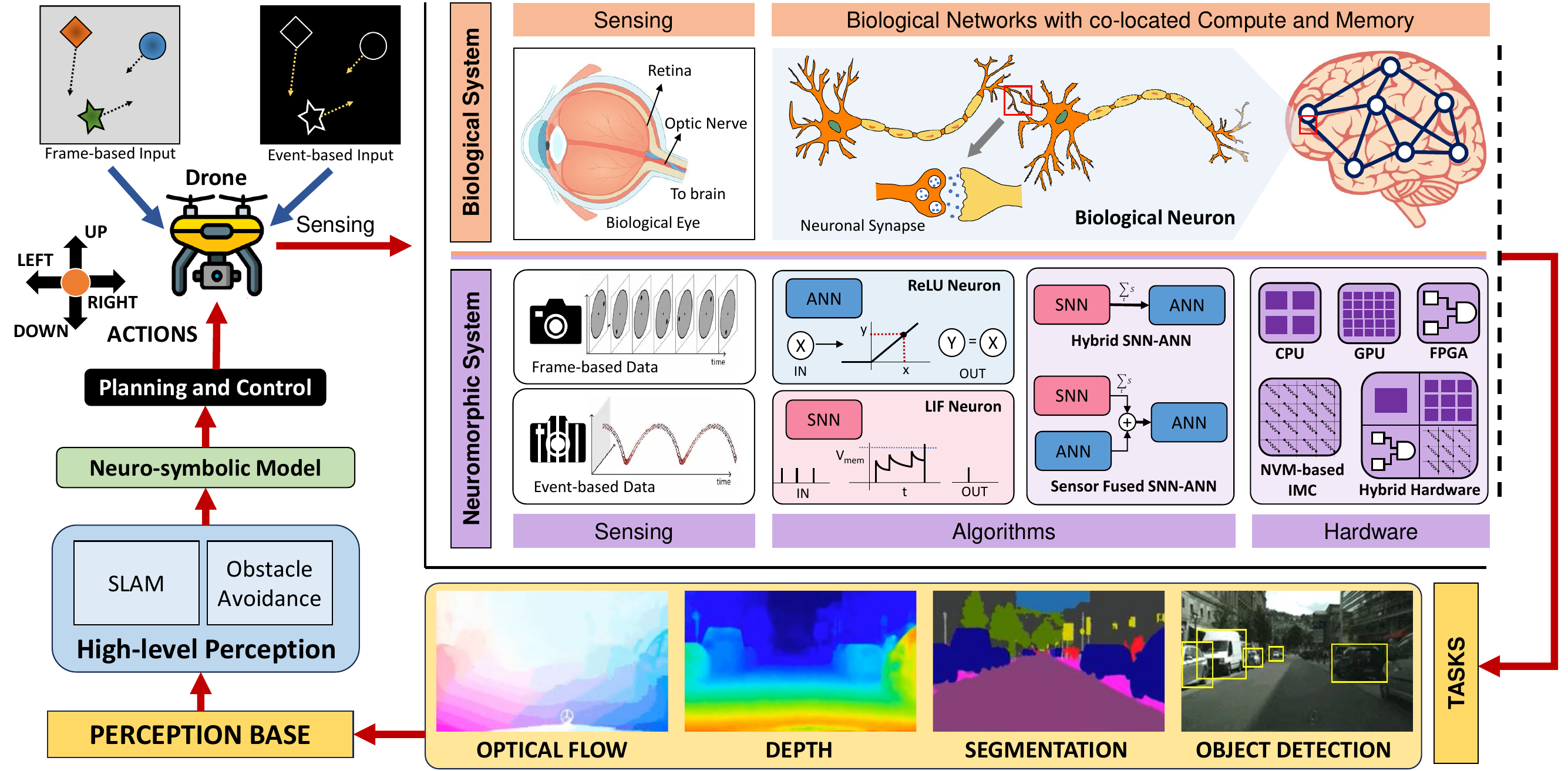}
\caption{\textbf{An end-to-end computing pipeline comparison sensing-processing-action loop between a biological and a neuromorphic system.} In a biological system, inputs are perceived as changes in intensity (events and frames) and color (frames) by the eye. In contrast, a neuromorphic system uses frame cameras to capture analog intensity at low rates and event cameras to detect motion-induced variations, generating events. The brain's parallel and recurrent connections enable computation within memory. Neuromorphic system emulates this by combining ANNs, SNNs, and hybrid ANN-SNN models to balance accuracy and efficiency. These algorithms also benefit from hardware acceleration via in-memory (IMC) and near-memory (NMC) computing by efficiently implementing synaptic functionality and, work alongside CPU/GPU architectures to enhance efficiency and reduce latency.}
  \label{fig:neuromorphic_vdn}
\end{figure*}

Despite these challenges, sensing-to-action loops can also offer significant and untapped opportunities for innovation and optimization. In the pursuit of these opportunities, in this paper we explore several fundamental questions. For example, in dynamic environments rich with features and high-dimensional stimuli, a key question is: \textit{Which sensory inputs are truly critical for decision-making?} Some inputs may overlap with existing model knowledge or provide redundant information, while others may have minimal impact on the task at hand. For example, in autonomous vehicles, repeated detection of static objects like buildings may add little new information, whereas detecting unexpected moving objects is crucial for safety. By selectively sensing novel or task-relevant features, systems can simplify processing and improve efficiency. To this end, in Sec. 3, we discuss generative sensing, where generative models reconstruct or ``dream out" most of the environment based on prior knowledge, thus reducing the need for exhaustive real-time data collection, without impacts on action abilities.

Action-to-sensing pathways present an equally compelling and unique avenue for enhancing system efficiency and robustness. A key question in this context is: \textit{How can control actions be used to proactively guide sensing, ensuring that data acquisition remains task-relevant and resource-efficient?} For instance, in a robotic navigation task, adjusting the sensor's field of view based on control objectives—such as steering toward a target—can reduce redundant data collection. Instead of passively gathering information, action-to-sensing frameworks can dynamically adjust sensing parameters, such as sampling rates and resolutions, to align with control demands. Towards this, as an exemplary technique, we discuss that Koopman operator-based representations provide a powerful approach by transforming complex non-linear dynamics into a linearly decomposable embedding space. By identifying key system eigenvalues, this framework can enable more efficient, task-informed control with fewer interactions, making autonomous systems more adaptable and resource-aware.

Another critical question is: \textit{How to ensure the reliability of sensing-to-action loops} amid evolving environmental dynamics, shifting application objectives, hardware degradations, and adversities, as sensing outputs directly influence actions, which, in turn, shape subsequent sensing stages. Without proper monitoring, these loops can drift from expected behaviors or become destabilized over time. For example, a misclassification in a surveillance drone's early detection phase could trigger inappropriate flight adjustments, skewing subsequent sensor coverage and compounding errors further. We examine methodologies that leverage robust statistical representations of intermediate features to detect deviations from the expected.

In multi-agent systems, sensing-to-action loops offer further opportunities for cooperative optimization and robust decision-making. A fundamental question in this context is: \textit{How can distributed agents coordinate sensing and actions to enhance global performance while balancing individual resource constraints?} In dynamic, feature-rich environments, redundant observations by multiple agents can lead to inefficient data processing, while critical, unique sensory inputs may remain underutilized. For example, in autonomous drone swarms, overlapping views of static objects may add little new information, whereas coordinated sensing of unexpected, fast-moving obstacles can improve safety and responsiveness.

Finally, we ask \textit{are there alternative representations that naturally fuse sensing, processing, and action variables} to enable systematic approaches that harness this unification for disruptive efficiency advancements? To this end, neuromorphic computing offers a compelling solution by integrating sensing and computation through event-driven, parallel processing, as illustrated in Fig.~\ref{fig:neuromorphic_vdn}. Unlike traditional clock-based systems, neuromorphic architectures trigger computations only in response to sensory events, enabling ultra-low-power, real-time performance. Neuromorphic computing also aligns seamlessly with the cyclical nature of sensing-to-action loops, providing a unified and efficient pathway through the tighter coupling of sensing, learning, and acting.

Systematically exploring each of these opportunities, Sec. 3 explores strategies for optimizing sensing-to-action loops and Sec. 4 discusses action-to-sensing loop optimization. Sec. 5 focuses on dynamically monitoring the fidelity of sensing-to-action loops. Sec. 6 examines neuromorphic representations of sensing, feature, and action variables. Sec. 7 extends the discussion to multi-agent interactions. Finally, Sec. 8 concludes.

\section{Optimizing ``Sensing-to-Action'' Pathways}

\begin{figure}[t!]
    \centering
    \includegraphics[width=\linewidth]{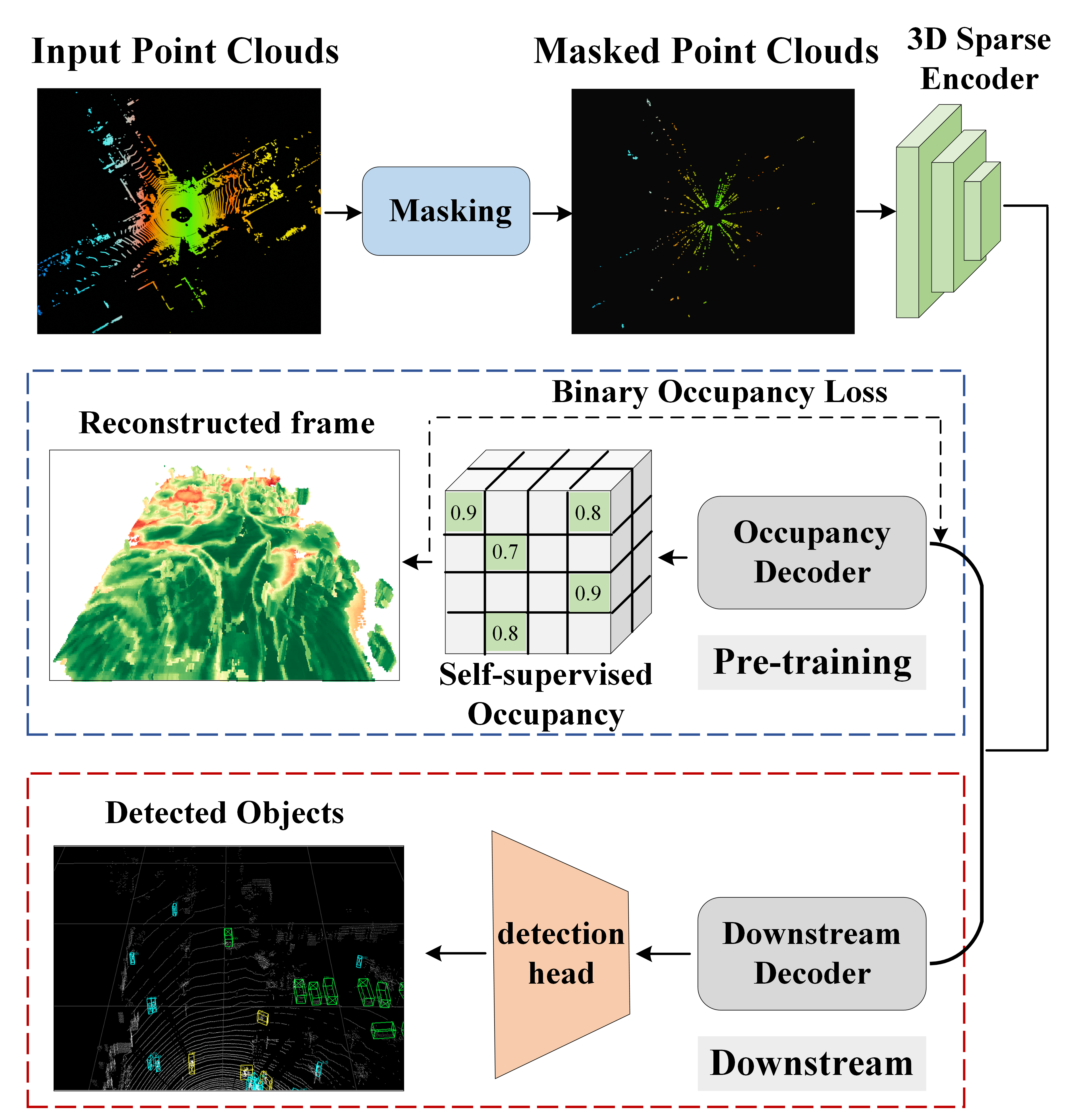}
\caption{\textbf{Generative Sensing: \textit{Sense only what you really need}:} Generative sensing optimizes resource use by focusing on essential environmental features, reducing unnecessary data collection and enhancing real-time responsiveness. For LiDAR proessing, in this approach, the input point cloud is voxelized and radially masked based on voxel distance from the sensor to minimize redundant information. A 3D spatially sparse convolutional encoder extracts latent features, while a decoder reconstructs the 3D scene, enabling efficient perception that supports adaptive sensing-to-action strategies.\label{network}}
\end{figure}

We begin by discussing the first optimization opportunity: prioritizing sensory inputs to balance observability and energy efficiency. A key question is \textit{which sensory inputs and domain regions are truly critical for decision-making?} Not all inputs contribute equally—some overlap with existing model knowledge or provide redundant information, while others have minimal impact on task outcomes. By selectively acquiring only the most relevant sensory data, active sensing systems can significantly reduce energy consumption without compromising performance, enabling efficient, real-time perception that adapts dynamically to environmental and task-specific demands. This approach is particularly impactful for active sensing modalities like LiDAR, which are essential for depth perception and robust object detection but have high energy consumption—around 25W compared to 1–2W for conventional cameras \cite{rablau2019lidar, sahin2019long}—posing significant challenges for resource-constrained edge systems.

Our recent research proposed the concept of \textit{generative sensing} \cite{tayebati2024sense}, which reimagines LiDAR-environment interaction by sampling only 8--10\% of the scene and using generative models to reconstruct unobserved regions. This builds on the insight that many scene regions are either predictable from pre-training or have minimal accuracy impact. In Fig. \ref{network}, such generative sensing is enabled by Radially Masked Autoencoding (R-MAE), which combines a masked autoencoder with a generative decoder for reconstructing unobserved regions and downstream object detection heads. 

\begin{table}[t]
\centering
\caption{Average Precision (AP) of R-MAE against current methods on KITTI (\textsuperscript{\dag} results are reproduced by us).}
\label{kitti-val-40}
\footnotesize{
\setlength{\tabcolsep}{10pt} 
\begin{tabular}{l|c|c|c}
\toprule
\textbf{Model} & \textbf{Car} & \textbf{Pedestrian} & \textbf{Cyclist} \\
\midrule

SECOND\textsuperscript{\dag} \cite{yan2018second} & 79.08 & 44.52 & 64.49 \\

+ Occ.-MAE\textsuperscript{\dag} \cite{min2023occupancy} & \underline{79.12$^{\textcolor{teal}{+0.04}}$} & 45.35$^{\textcolor{teal}{+0.83}}$ & 63.27$^{\textcolor{teal}{-1.22}}$ \\

+ ALSO\textsuperscript{\dag} \cite{boulch2023also} & 78.98$^{\textcolor{teal}{-0.10}}$ & 45.33$^{\textcolor{teal}{+0.81}}$ & 66.53$^{\textcolor{teal}{+2.04}}$ \\

+ R-MAE (Ours) & \textbf{79.10$^{\textcolor{teal}{+0.02}}$} & \underline{\textbf{46.93$^{\textcolor{teal}{+2.41}}$}} & \underline{\textbf{67.75$^{\textcolor{teal}{+3.26}}$}} \\

\midrule
PV-RCNN \cite{shi2020pv} & 82.28 & 51.51 & 69.45 \\

+ Occ.-MAE\textsuperscript{\dag} \cite{min2023occupancy} & 82.43$^{\textcolor{teal}{+0.15}}$ & 48.13$^{\textcolor{teal}{-3.38}}$ & 71.51$^{\textcolor{teal}{+2.06}}$ \\

+ ALSO\textsuperscript{\dag} \cite{boulch2023also} & 82.52$^{\textcolor{teal}{+0.24}}$ & \underline{52.63$^{\textcolor{teal}{+1.12}}$} & 70.20$^{\textcolor{teal}{+0.75}}$ \\

+ R-MAE (Ours) & \underline{\textbf{82.82$^{\textcolor{teal}{+0.54}}$}} & \textbf{51.61$^{\textcolor{teal}{+0.10}}$} & \underline{\textbf{73.82$^{\textcolor{teal}{+4.37}}$}} \\

\bottomrule
\end{tabular}
}
\end{table}

\begin{table}[t]
\centering
\caption{Comparison of Conventional LiDAR and R-MAE Framework}
\footnotesize
\setlength{\tabcolsep}{10pt} 
\begin{tabular}{l|c|c}
\hline
\textbf{Metric}             & \textbf{Conventional} &  \textbf{R-MAE}   \\ \hline
Scene Coverage              & 100\% (full scan)    & $<10\%$ (active) \\ 
Energy per Laser Pulse      & 50 µJ                & 5.5 µJ           \\ 
Model Parameters            & Not applicable       & 830K             \\ 
FLOPs per 360° Scan         & None                 & 335M             \\ \hline
Sensing Energy per Scan     & 72 mJ                & 792 µJ           \\ 
Reconstruction Overhead     & Not applicable       & 7.1 mJ           \\ \hline
\end{tabular}
\label{tab:gensense-energy}
\end{table}

R-MAE uses a range-aware radial masking strategy to optimize LiDAR beam activation while accounting for light propagation physics. The masking operates in two stages: (1) grouping voxels into angular segments and sampling a subset for sensing, and (2) applying distance-dependent probabilistic masking to address the $R^4$ energy scaling with range. This approach addresses LiDAR’s energy-accuracy-range trade-offs without hardware modifications. While improving angular precision ($\Delta\theta$) typically requires increasing the aperture diameter ($D$) or using shorter wavelengths ($\lambda$), practical constraints like size and eye safety limit these options. Instead, R-MAE’s two-stage masking reduces redundant data collection and conserves energy, maintaining robust scene coverage within constraints.

R-MAE architecture, in Fig. \ref{network}, integrates key components: a 3D sparse convolutional encoder processes the partial point cloud into a latent representation capturing geometric and semantic features, followed by an occupancy decoder that reconstructs the full 3D scene. To balance efficiency and reconstruction quality, the encoder processes only non-empty voxels, preserving geometric structure while reducing memory usage compared to Transformer-based methods \cite{spconv2022, xu2023mv, hess2023masked}. The decoder uses deconvolution layers with batch normalization and ReLU activation to progressively refine the scene at higher resolutions, with a binary cross-entropy loss ensuring accurate occupancy prediction and spatial consistency.

Our prior work \cite{tayebati2024sense} demonstrated the effectiveness of R-MAE through extensive evaluations across multiple datasets. As shown in \autoref{kitti-val-40}, R-MAE exhibited strong generalization, significantly improving accuracy on the KITTI validation set (40 recall positions at moderate difficulty). On the Waymo dataset, it outperformed baselines by up to 5.59\% in mAP/mAPH, even with 90\% of the scene masked. Similar gains were observed on the nuScenes dataset, improving LiDAR-only models' NDS scores by 2.31\% to 3.17\%. Additionally, R-MAE achieved substantial energy savings, reducing average laser pulse energy to 5.5 µJ compared to 50 µJ in conventional systems. While adding computational overhead ($\sim$830K parameters and $\sim$335M FLOPs per 360° scan), the combined energy consumption of sensing and reconstruction was 9.11$\times$ lower than traditional LiDAR, particularly for long-range measurements where energy costs scale with the fourth power of distance.

Generative sensing thus represents a significant advancement for ultra-frugal perception systems, rethinking the sensing-processing pipeline to leverage trends in sensing and computing energy costs. This approach can extend beyond LiDAR to modalities such as radar, cameras, and acoustic sensors, where selective sampling and reconstruction similarly enhance performance and energy efficiency. Future work could explore adaptive masking, multi-modal fusion, and broader applications to active sensing, advancing sensor data compression and reconstruction for ultra-low-power autonomous systems. To minimize generative sensing workload techniques such as in-memory computing \cite{shukla2020mc,shukla2021ultralow,nasrin2021mf}, analog computing \cite{shylendra2020low}, and beyond CMOS devices \cite{trivedi2014ultra, sangwan2022two,finocchio2024roadmap} can be explored.

\section{Optimizing ``Action-to-Sensing'' Pathways}

\begin{figure}
    \centering
    \includegraphics[width = \linewidth]{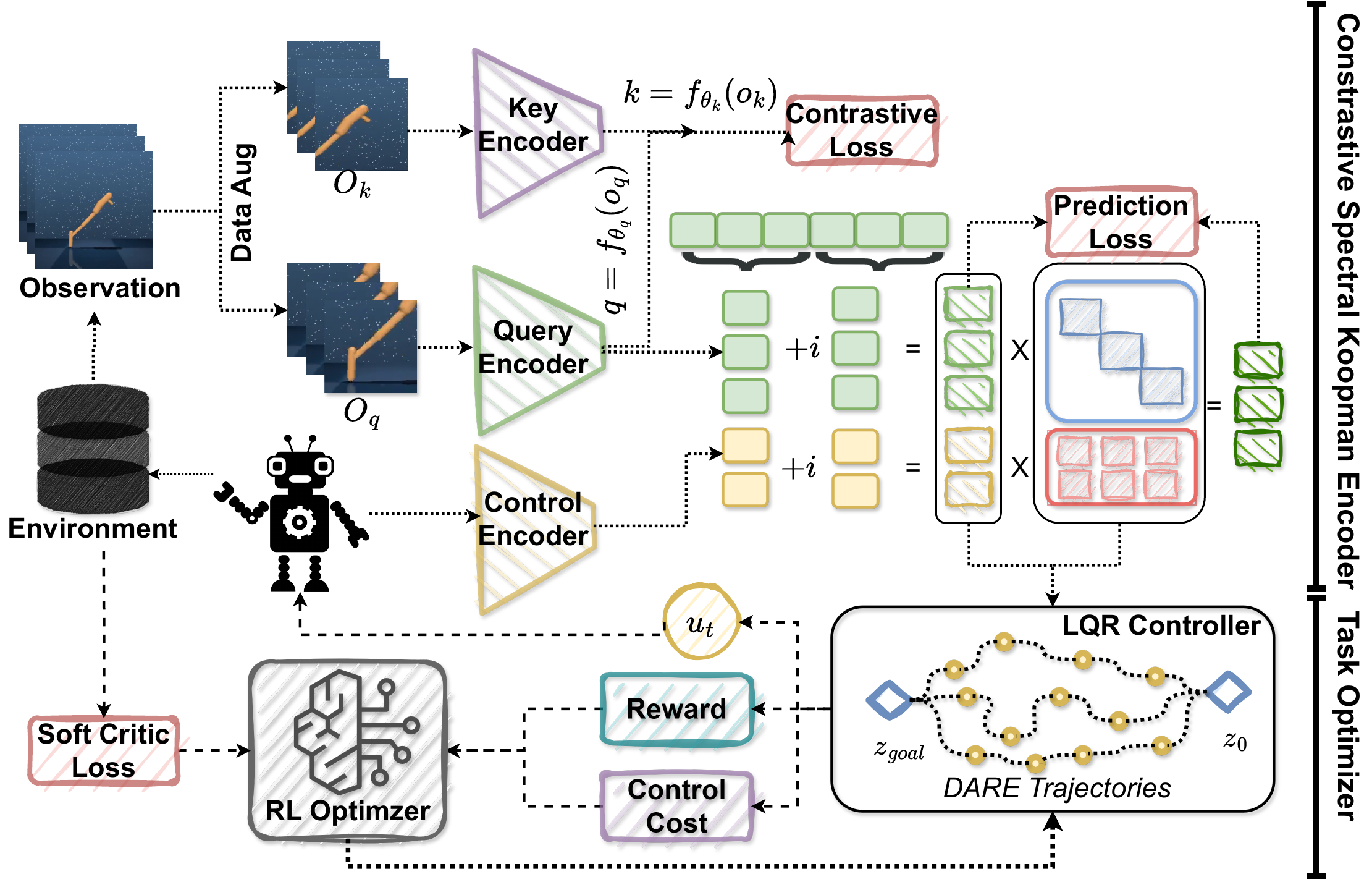}
    \caption{Our approach conditions visual representations on the task policy by incorporating contrastive spectral Koopman encoding and reinforcement learning (RL)-guided control. This high-level framework unifies perception and control, enabling task-aware sensing adjustments. The RoboKoop model leverages these representations to dynamically adjust sensing parameters based on control objectives. (Adapted from RoboKoop\cite{kumawat2024robokoopefficientcontrolconditioned})}
    \label{fig:main_2}
\end{figure}

While sensing-to-action loops focus on extracting sensory data to inform decisions, the reverse pathway—action-to-sensing—presents an equally important optimization opportunity: using system actions to guide and refine subsequent sensing. Autonomous systems often operate in dynamic environments \cite{9892390,10546823, samal2} where sensory context \cite{radar_guided} and task priorities \cite{kumawat2024stage, pmlr-v242-kumawat24a} evolve, raising a key research question: \textit{How can actions be leveraged to proactively adjust sensing strategies to improve efficiency without compromising situational awareness?} By incorporating feedback from past actions and outcomes, action-to-sensing pathways can dynamically modulate sensing parameters such as focus, sampling rates \cite{kumawat2024adacredadaptivecausaldecision}, and resolution based on the current context. This transforms sensing from passive data collection into an active, goal-directed process that adapts to environmental changes and resource constraints.

\begin{figure}
    \centering
    \begin{subfigure}{0.48\textwidth}
        \centering
        \includegraphics[width=\linewidth]{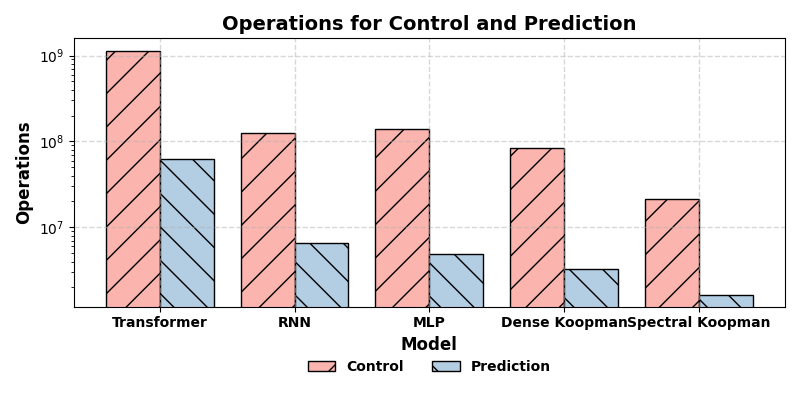}
    \end{subfigure}\hfill
    \begin{subfigure}{0.48\textwidth}
        \centering
        \includegraphics[width=\linewidth]{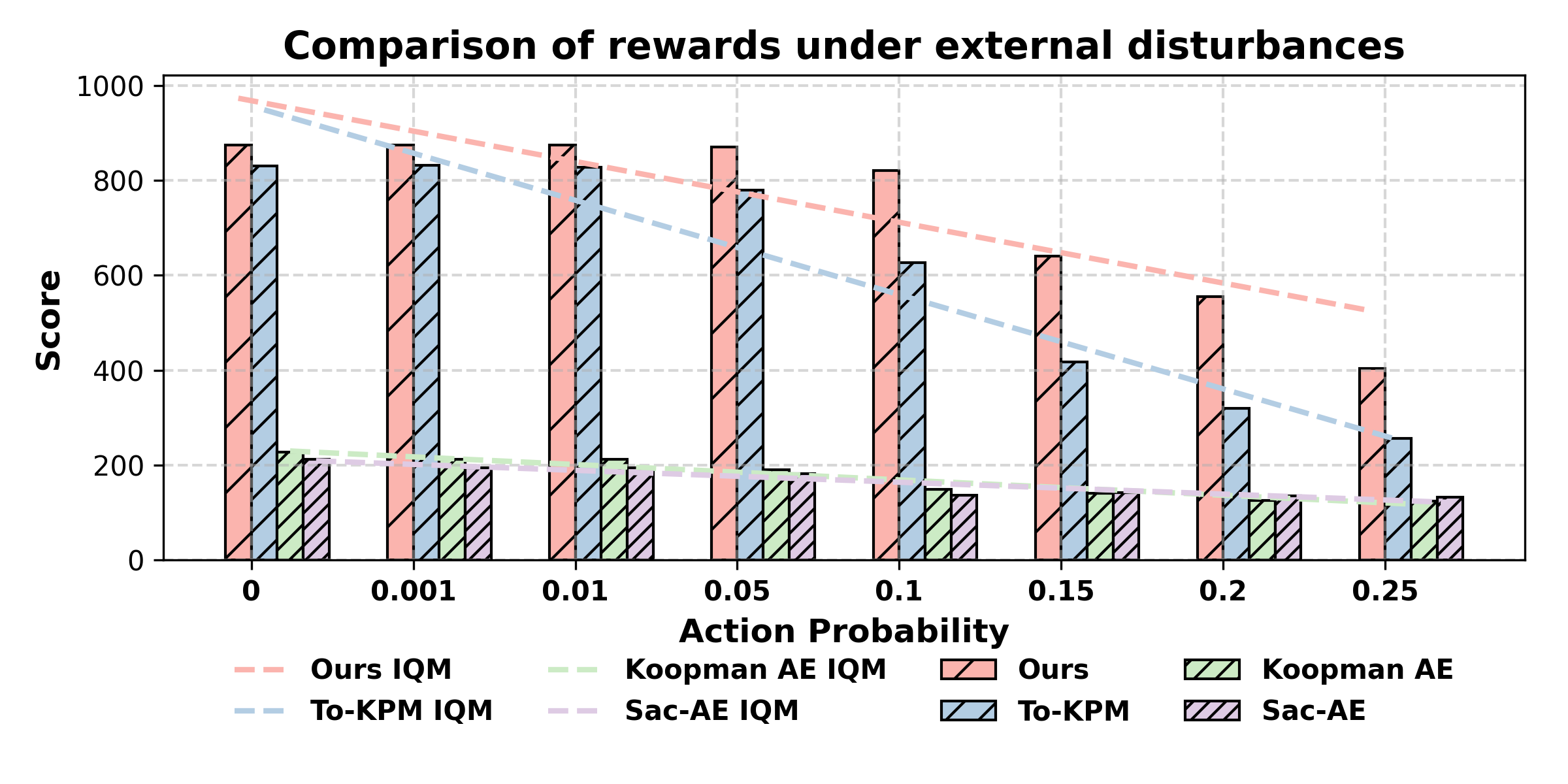}
    \end{subfigure}
     \caption{(a) Computational load of state-of-the-art dynamical models. (b) Performance under external disturbances. (Adapted from RoboKoop\cite{kumawat2024robokoopefficientcontrolconditioned})}
     \label{fig:action-prob-plot}
\end{figure}

Towards this goal, our prior work \cite{kumawat2024robokoopefficientcontrolconditioned} hypothesized that robust agent representations can be learned with fewer interactions if the task embedding space can be modeled linearly and a finite set of stable (negative) eigenvalues of the Koopman operator are identified. Fig.~\ref{fig:main_2} illustrates the approach for learning a linear Koopman embedding manifold using a contrastive spectral Koopman encoder. This encoder generates key and query samples for each observation at time \(t\), where positive samples apply random cropping augmentations to the state \(x_t\), and negative samples use augmentations on other states. The query encoder maps visual observations to a complex-valued Koopman embedding space with learnable eigenvalues \(\mu_i + j\omega\).

Using this embedding and the spectral Koopman operator, optimal control strategies are derived by solving a Linear Quadratic Regulator (LQR) problem over a finite time horizon. The goal state, provided in visual space, is similarly mapped via the key encoder. Dual Q-value functions within the Soft Actor-Critic (SAC) framework guide updates based on the LQR controller's cost. Training involves optimizing three key losses: the SAC loss for training the critic and Koopman parameters, the contrastive loss to refine the encoder, and the next latent prediction loss to regularize Koopman embedding dynamics.

We assessed computational efficiency by implementing MLP-based dynamics \cite{srini}, a dense Koopman model \cite{lyu2023taskoriented}, a Transformer model \cite{vaswani2023attention, chen2021decision_trans}, and a recurrent model \cite{hafner2020dream}. Fig.~\ref{fig:action-prob-plot}(a) shows that our spectral Koopman-based approach required the fewest Multiply-Accumulate (MAC) operations for control and prediction, highlighting its efficiency in dynamic system modeling. To test robustness, we applied an external force \(F \sim \text{Uniform}(a_{\text{min}}, a_{\text{max}})\) to the cart-pole system during evaluation, with a disturbance probability \(p\). Fig.~\ref{fig:action-prob-plot}(b) shows that our model maintained high performance even with a disturbance probability of 0.25, demonstrating superior resilience compared to other methods.

This analysis shows that action-to-sensing pathways, combined with efficient Koopman-based representations, enhance the adaptability and resilience of autonomous systems by linking control actions directly to optimized sensing strategies. Future work could extend this framework to handle non-stationary dynamics by learning time-varying Koopman operators that adapt to environmental shifts, such as sensor degradation or task transitions. Additionally, incorporating uncertainty quantification within Koopman representations to adjust sensing actions based on confidence estimates can help reduce cascading errors in uncertain environments.

\begin{figure}[t]
    \centering
    \includegraphics[width=\linewidth,clip]{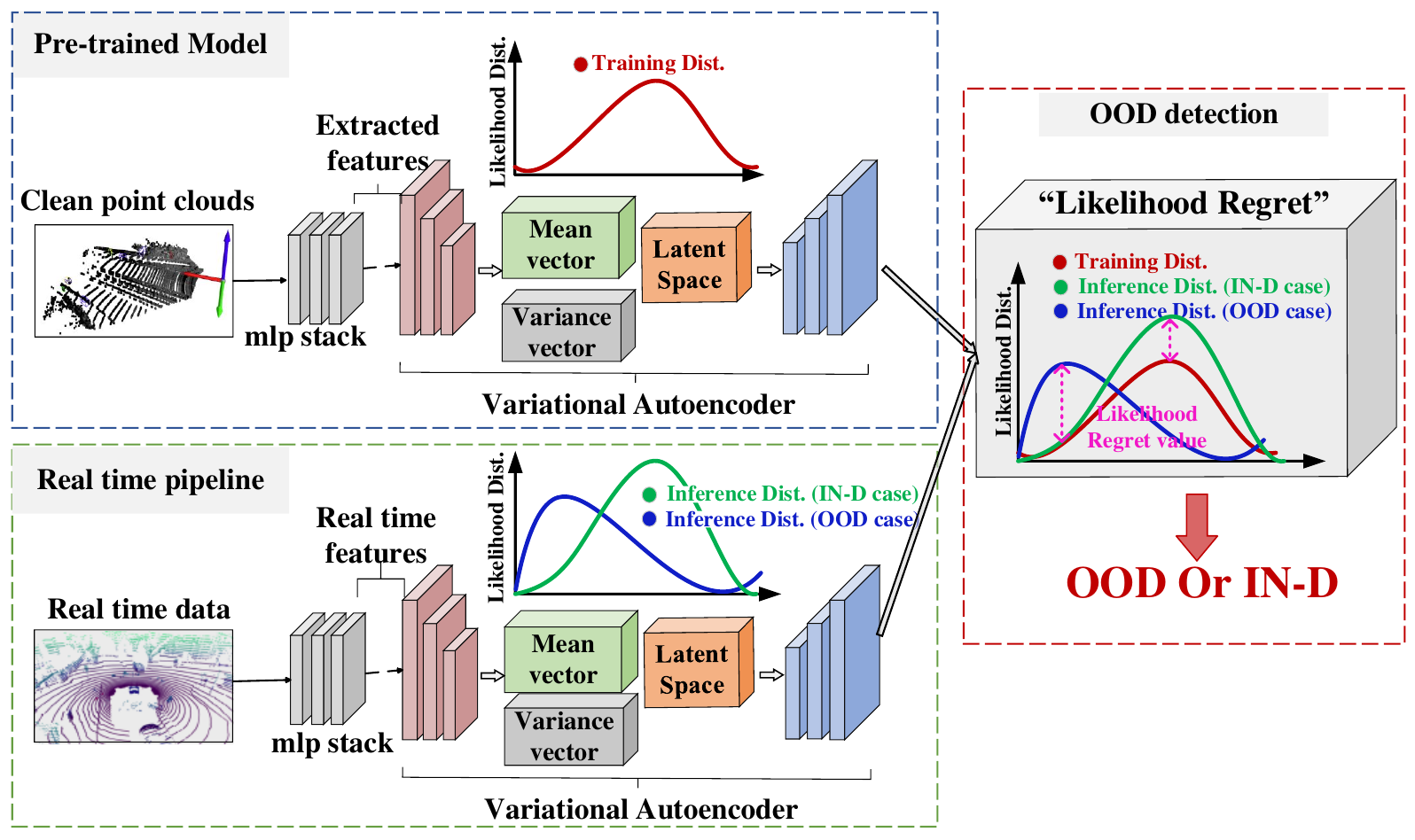}
\caption{\textbf{Ensuring Sensing-Action Loop Reliability:} STARNet enhances the reliability of sensing-to-action loops by ingesting feature representations from primary task networks. A VAE models the typical distribution of these features, and during inference, STARNet uses gradient-free optimization to compute likelihood regret, identifying discrepancies between sensed and learned distributions to alert the system to potential inaccuracies.}
    \label{fig: STARNET}
\end{figure}

\section{Reliability of Sensing-to-Action Loops}
A key challenge is maintaining accurate and consistent feedback as systems interact with evolving surroundings. This raises critical questions: \textit{How can sensing-to-action loops remain stable despite environmental changes, sensor degradation, or adversarial disruptions? What mechanisms can detect and correct deviations before they cascade into failures?} While adaptive, task-conditioned mechanisms in previous sections enhance flexibility and responsiveness, they also introduce risks—dynamic adjustments may amplify noise, propagate erroneous data, or destabilize feedback loops\cite{heinzler2019weather, gallego2020event, sole2004solid}. Without robust monitoring, these systems risk propagating inaccuracies.

To address these challenges, we proposed STARNet, a two-stage mechanism that monitors sensor data trustworthiness in real time to maintain sensing-to-action loop integrity \cite{darabi2023starnet}. Instead of relying on static sensing configurations, STARNet evaluates intermediate sensor features from primary tasks to detect untrustworthy data streams deviating from expected distributions. As shown in Fig.~\ref{fig: STARNET}, STARNet’s core components include a Variational Autoencoder (VAE) that learns the distribution of normal sensor embeddings to model complex, high-dimensional data. A Likelihood Regret (LR) metric \cite{xiao2020likelihood} quantifies how much the VAE's distribution must adjust for a new input, with large LR scores indicating anomalies. To reduce computational overhead, STARNet uses gradient-free optimization techniques such as Simultaneous Perturbation Stochastic Approximation (SPSA) \cite{ghadimi2013stochastic, bhatnagar2023generalized}, making it suitable for low-power edge devices. Additionally, Low-Rank Adaptation (LoRA) \cite{hu2021lora, jayasuriya2024neural} enables efficient on-device fine-tuning by constraining updates to a low-dimensional subspace while preserving core model weights for fast adaptation.

\begin{figure}[t!]
    \centering
    \includegraphics[width=\linewidth]{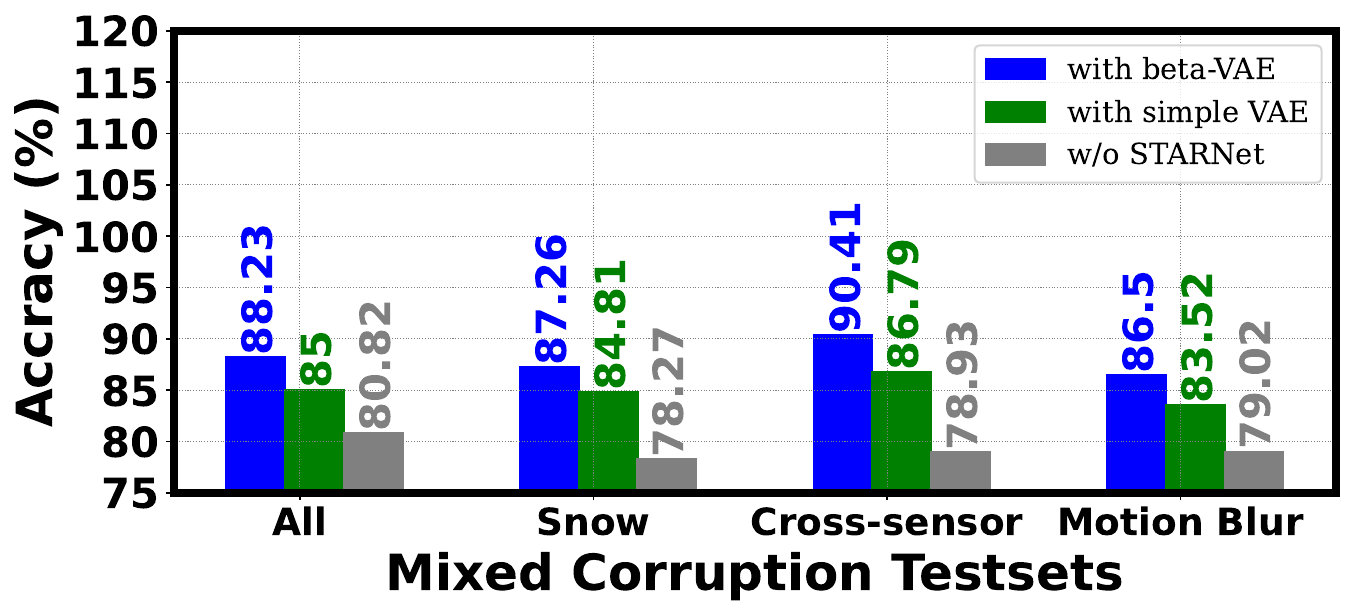}    
    \caption{\textbf{Object Detection Accuracy for KITTI Dataset:} The VAE-based approach analyzes LiDAR point clouds and object labels, producing bounding boxes for cars, pedestrians, and cyclists. The network was tested under challenging conditions, such as varying snow intensities and other corruptions.}
    \label{fig:ACC}
\end{figure}

We evaluated STARNet using KITTI-C data \cite{kong2023robo3d} across natural corruptions (e.g., rain, fog) \cite{heinzler2019weather}, external disruptions (e.g., beam missing, motion blur) \cite{gandhi2007pedestrian}, and internal sensor failures (e.g., crosstalk  \cite{diehm2018mitigation}, cross-sensor interference \cite{kalasapati2022robustness}). In LiDAR-only tests, STARNet achieved AUC values above 0.90 for crosstalk (0.9658) and cross-sensor interference (0.9938), demonstrating strong detection capabilities without explicit training on these faults. When fusing LiDAR with camera inputs, STARNet further improved anomaly detection under heavy snow while maintaining high task accuracy for detecting cars and pedestrians by filtering unreliable sensor data. As shown in Fig.~\ref{fig:ACC}, STARNet increased object detection accuracy by $\sim$15\%, restoring performance to clean data.

By reinforcing sensing-to-action loops with proactive anomaly detection, STARNet ensures reliable adaptive sensing and robust decision-making in complex environments. Future enhancements include context-aware anomaly detection to reduce false positives, adaptive fusion to adjust sensor weights based on reliability, and temporal consistency checks for detecting gradual sensor degradation. Additionally, uncertainty-aware \cite{stutts2024mutual,stutts2023lightweight} control mechanisms can modulate actions based on confidence levels \cite{darabi2024navigating, parente2024conformalized}.

\begin{figure*}[t]
  \centering
\includegraphics[width=\linewidth]{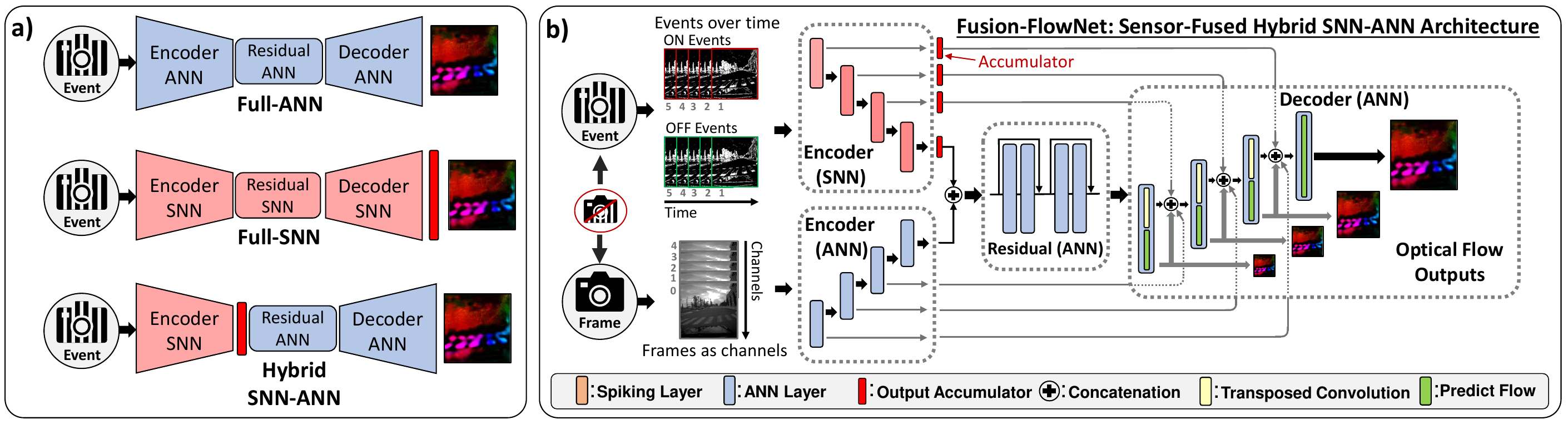}
   \caption{\textbf{Neuromorphic sensing-action loop architectures.} a) Full-ANN~\cite{zhu2018ev}, Full-SNN~\cite{kosta2022adaptive}, and Hybrid SNN-ANN~\cite{lee2020spike} models for optical flow estimation using event data. b) Fusion-FlowNet~\cite{lee2022fusion} integrates event-based and frame-based modalities for enhanced feature extraction. Outputs are aggregated at the final SNN layer. (Adapted from Fusion-FlowNet~\cite{lee2022fusion}).}
  \label{fig:opflow}
  \vspace{-3mm}
\end{figure*}

\section{Sensing-Action Loops in Neuromorphic Systems}
Biological systems, renowned for their efficiency in sensing, processing, and interacting with their environment, inspire the design of sensing-to-action frameworks for resource-constrained edge devices. For instance, the fruit fly (Drosophila melanogaster) navigates complex environments using just 100,000 neurons, consuming only 26.6 W/kg during flight~\cite{zhu2020kinematics}. This efficiency arises from an integrated architecture where neurons simultaneously perform sensing, computation, and memory—forming tightly coupled sensing-action loops that selectively respond to relevant stimuli while remaining inactive otherwise. \textit{What makes biological sensing-action loops resilient, and how can these principles inform artificial systems?}

In our prior work~\cite{rathi2023exploring}, we explored neuromorphic sensing-action loops inspired by event-driven, bio-plausible processing to create asynchronous architectures that compute only for meaningful events. Neuromorphic systems \cite{seo2024random} also co-locate computation and memory, enabling massively parallel processing while reducing energy consumption and data transfer delays—making them ideal for edge systems. Comprehensive neuromorphic frameworks leverage event-based sensors, spiking neural networks (SNNs), and bio-inspired learning to emulate the brain's sensing-processing-action loop. This approach reduces latency, conserves energy, and supports adaptive control, bringing artificial an biological systems closer.

Recent advances in event-driven sensors have further strengthened the potential of neuromorphic sensing-action loops. Frame-based cameras, while standard for vision tasks, are unsuitable during rapid motion scenario due to their low temporal resolution, increasing storage and latency. By contrast, event-based cameras like DVS128~\cite{dvs128} and DAVIS240~\cite{davis240} asynchronously capture pixel-wise intensity changes offering superior temporal resolution ($10 \mu s$ vs $3 ms$), lower power consumption ($10 mW$ vs $3 W$), and wider dynamic range ($120 dB$ vs $74 dB$) compared to frame-based cameras~\cite{gallego2020event}. However, frame-based data still enhances accuracy in some tasks~\cite{messikommer2023data}, highlighting the need for sensor fusion approaches~\cite{lee2022fusion}.

SNNs~\cite{maass1997networks} are well-suited for processing event-based sensor inputs due to their sparse, event-driven computations and intrinsic memory for sequential tasks~\cite{roy2019towards}. This makes them efficient alternatives to RNNs and LSTMs~\cite{ponghiran2022spiking}. However, training deep SNNs is challenging due to vanishing spikes and non-differentiable activations~\cite{rathi2020enabling, neftci2019surrogate}. Recent advancements, such as ANN-to-SNN conversion~\cite{cao2015spiking}, learnable neuronal dynamics~\cite{rathi2021diet, kosta2022adaptive}, and surrogate gradient methods~\cite{neftci2019surrogate, lee2020enabling}, address these limitations. For example, a recent work, Adaptive-SpikeNet~\cite{kosta2022adaptive} employs learnable spiking neuronal dynamics to achieve 20\% lower average endpoint error (AEE) than traditional ANNs for optical flow estimation, with 48$\times$ fewer parameters and consuming 10$\times$ less energy.

\begin{figure}[t!]
  \centering
\includegraphics[width=\linewidth]{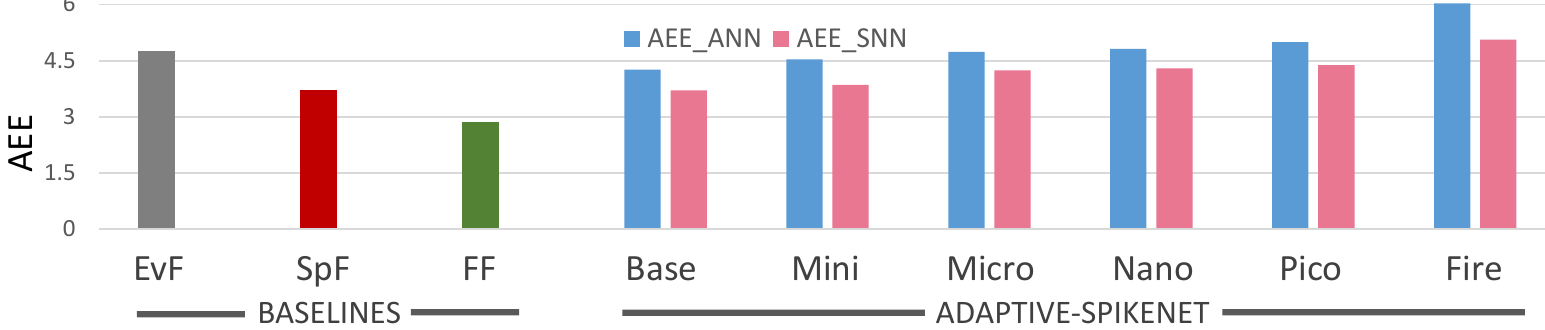}
   \caption{\textbf{Average Endpoint Error (AEE) comparison for Optical flow estimation on the MVSEC~\cite{zhu2018multivehicle} dataset.} The left shows the Average Endpoint Error (AEE) for baseline models, EvFlow-Net (EvF)~\cite{zhu2018ev}, Spike-FlowNet (SpF)~\cite{lee2020spike}, and Fusion-FlowNet (FF)~\cite{lee2022fusion}). The right showcases how AEE varies with model size for Adaptive-SpikeNet and corresponding full-ANN models. (Adapted from Adaptive-SpikeNet~\cite{kosta2022adaptive}).}
  \label{fig:comparison}
\end{figure}

Efforts to develop hybrid SNN-ANN architectures leverage the strengths of both networks, improving performance while reducing training complexity and energy consumption. Spike-FlowNet~\cite{lee2020spike}, which combines an SNN encoder with an ANN decoder for optical flow estimation, outperforms full-ANN models~\cite{zhu2018ev} on the MVSEC dataset~\cite{zhu2018multivehicle} with a 1.21$\times$ energy reduction. Sensor-fusion models like Fusion-FlowNet~\cite{lee2022fusion} integrate events and frames, achieving 40\% lower error with nearly half the parameters and 1.87$\times$ lower energy. For simpler tasks like object detection, full-SNN models excel—DOTIE~\cite{nagaraj2023dotie}, a lightweight, single-layer SNN, filters events based on speed and clusters them into bounding boxes. 

These algorithmic advancements when coupled with suitable neuromorphic hardware powered by in-memory computing (IMC) architectures such as~\cite{kim2022neuro, agrawal2021impulse, ankit2017resparc, lee2024highly, darabi2024adc}, could enable the sensing-processing-action loop characteristic of the brain, resulting in a truly end-to-end neuromorphic system capable of brain-like intelligence and efficiency.

\section{Federated, Multi-Agent Sensing-Action Loops}
Multi-agent sensing-action loops have the potential to enhance system-wide adaptability and efficiency by enabling agents to collaboratively sense, learn, and act. However, key research questions remain: \textit{How can agents dynamically share sensing and computation tasks to avoid redundancy while maintaining both individual and collective performance? How can robust decision-making be ensured in the face of network latency, hardware heterogeneity, and agent failures?} Addressing these challenges requires frameworks that balance collaboration and independence, adapt to real-time conditions, and optimize resource allocation across a distributed network of agents.

\begin{figure}[t!]
    \centering
    \includegraphics[width=\linewidth]{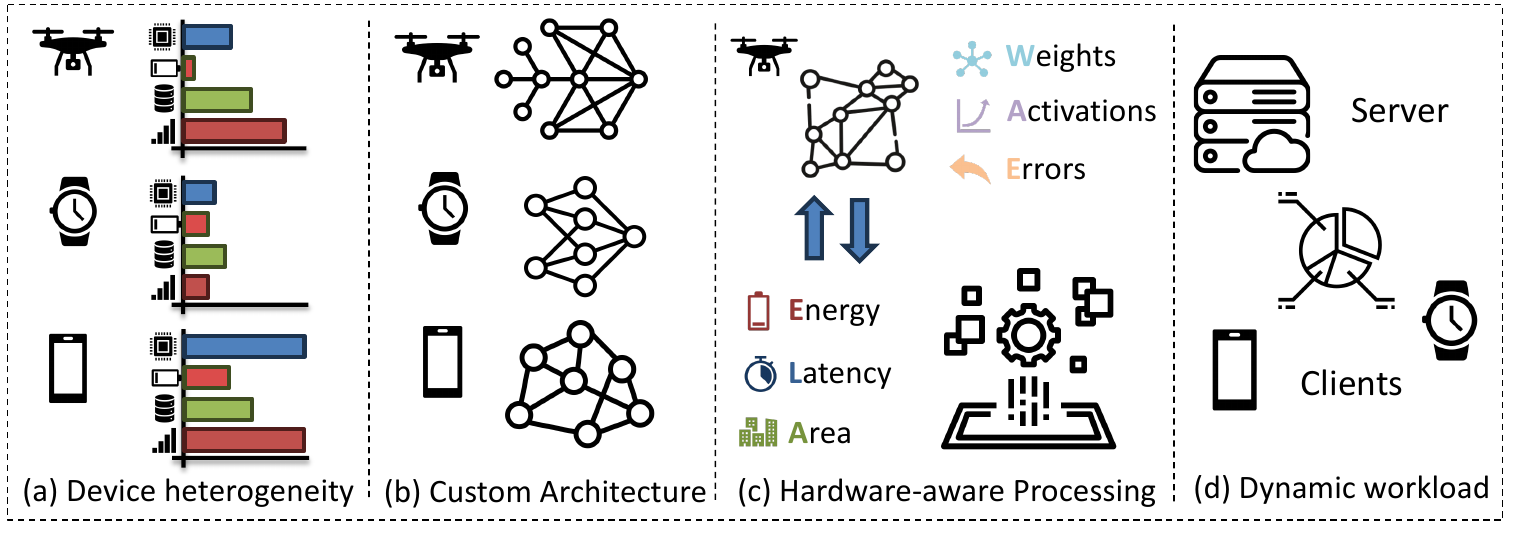}
    \caption{Key aspects of dynamic multi-agent systems: resource heterogeneity, adaptable architectures, hardware-aware optimization, and workload management across server-client interactions.}
    \label{fig:multi-agent}
\end{figure}

Federated learning (FL) has emerged as a promising approach to decentralized, collaborative learning without the need to share raw data. By aggregating insights from distributed agents, FL preserves data privacy and enhances security, making it effective for applications in healthcare \cite{antunes2022federated}, Internet of Things (IoT) \cite{khan2021federated}, and autonomous systems \cite{xianjia2021federated}. However, real-world FL deployments face challenges such as hardware heterogeneity, intermittent connectivity, and varying application requirements. Participating devices often have diverse constraints, including differences in compute power, memory, and energy availability (Fig.~\ref{fig:multi-agent}). Traditional FL approaches, which assume uniform client capabilities and static models, are ill-suited for such diverse, dynamic environments.

\begin{figure}[t!]
    \centering
    \includegraphics[width=0.8\linewidth]{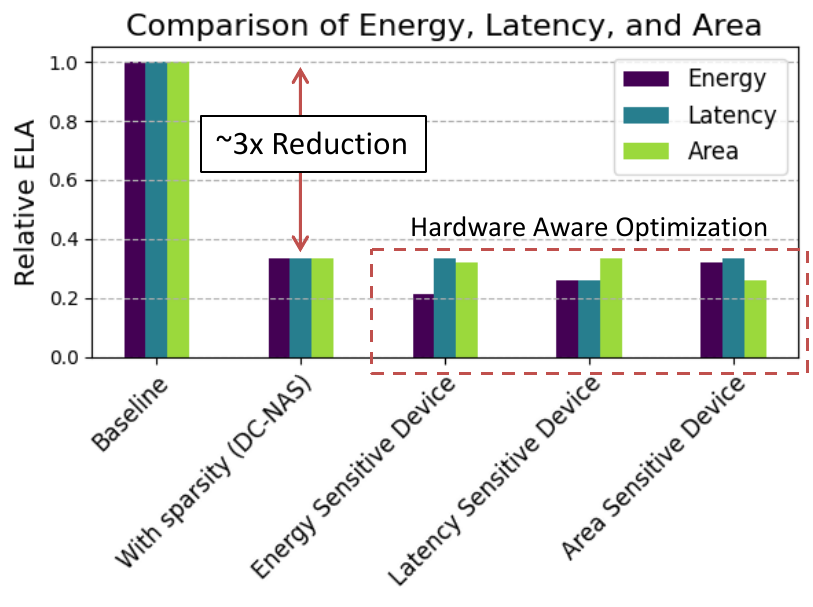}
    \caption{Performance comparison of DC-NAS and HaLo-FL on the CIFAR-10 dataset, showing relative reductions in energy, latency, and area with adaptive model optimization.}
    \label{fig:results-multi-device}
\end{figure}

Dynamic federated learning frameworks address this heterogeneity by adapting models and training processes in real time. DC-NAS \cite{VENKATESHA2023569}, for instance, tailors neural network architectures to client-specific constraints through topology and channel pruning, enabling efficient collaboration without overburdening resource-limited agents. By dynamically adjusting model complexity, DC-NAS improves both training efficiency and robustness. Similarly, HaLo-FL \cite{halo-fl} incorporates a hardware-aware precision selector that optimizes weights, activations, and gradients based on client capabilities, reducing energy consumption and latency while preserving accuracy. This adaptability is enabled by a precision-reconfigurable simulator, allowing real-time adjustments to meet energy, latency, and area constraints.

Speculative decoding \cite{leviathan2023fast} exemplifies how edge-cloud collaboration can further enhance multi-agent systems. By parallelizing token predictions and verifying outputs probabilistically, speculative decoding accelerates autoregressive tasks such as conversational AI and multimodal processing. This approach enables resource-constrained edge devices to perform lightweight inference locally while offloading complex computations to the cloud. For example, quadruped robots in disaster zones can process multimodal inputs—such as text instructions, visual data, and sensor readings—to generate context-aware responses in real time. The edge handles low-latency predictions, while the cloud refines and updates models as needed, reducing communication overhead even in dynamic scenarios.

Fig.~\ref{fig:results-multi-device} illustrates how adaptive frameworks like DC-NAS and HaLo-FL significantly reduce energy, latency, and area utilization while maintaining performance on datasets like CIFAR-10. These approaches highlight the importance of integrating adaptive model architectures, real-time profiling, and predictive resource allocation in distributed multi-agent systems. Emerging hardware paradigms, such as in-memory computing and low-precision representations, further support energy-efficient execution by minimizing data movement and overheads.

Ultimately, advancing multi-agent sensing-action loops requires bridging algorithmic innovations with hardware-aware design. This includes balancing edge-cloud workloads, synchronizing model updates, and leveraging speculative decoding for efficient decision-making. By co-optimizing sensing, computation, and communication, these systems can achieve robust, scalable performance, paving the way for responsive and resource-efficient AI across diverse applications.

\section{Conclusions}  
This article highlighted the crucial role of sensing-to-action loops in enabling real-time decision-making for autonomous edge computing. These loops can enhance system adaptability and resource efficiency by dynamically aligning sensor inputs with computational models for task-specific control. However, they also introduce challenges such as synchronization delays, resource constraints, and cascading errors, necessitating robust cross-layer co-design strategies. To address these challenges, we discussed generative sensing frameworks that selectively sense critical parts of the environment and use generative models to reconstruct predictable regions. This approach showed that only 8\% of the environment needs to be actively sensed, significantly reducing sensing overhead. Similarly, action-to-sensing pathways demonstrated how control objectives can proactively adjust sensing strategies to maintain situational awareness while minimizing redundant data acquisition. The use of Koopman operator-based representations improved computational efficiency across models, including transformers.

Despite these gains, sensing-to-action loops face destabilization risks due to runtime adaptations. To mitigate this, we presented the STAR-Net framework, which employs metrics like likelihood regret to monitor and enhance the reliability of these loops, improving prediction accuracy by over 10\% on complex datasets. We also demonstrated that multi-agent sensing-to-action loops, leveraging federated learning and distributed collaboration, can achieve a threefold reduction in energy consumption. Lastly, we explored efficient representations that integrate sensing, perception, and action, highlighting the potential of neuromorphic systems. These event-driven architectures synchronize sensing and computation, enabling energy-efficient, low-latency processing well-suited for resource-constrained environments.

\vspace{3pt}
\noindent\textbf{Acknowledgment:} This work was supported by COGNISENSE and CoCoSys, two of seven centers in JUMP 2.0, a Semiconductor Research Corporation (SRC) program sponsored by DARPA, and NSF Awards \#2329096, \#2106964, and \#2317974.

\bibliographystyle{IEEEtran}
\bibliography{main}
\end{document}